\documentclass[preprint,12pt]{elsarticle}

\usepackage[utf8]{inputenc}
\usepackage[T1]{fontenc}
\usepackage{lmodern}
\usepackage[english]{babel}
\usepackage{amsmath, amssymb, amsfonts, bbm}
\usepackage{graphicx}
\usepackage{booktabs}
\usepackage{geometry}
\usepackage{hyperref}
\usepackage{float}
\usepackage{titlesec}
\usepackage{microtype}
\usepackage{lipsum} % For dummy text if needed, though we will fill with content
\usepackage{caption}
\usepackage{lineno}
\usepackage{tikz}
\usetikzlibrary{positioning, fit, calc, shapes.geometric, arrows.meta, shadows, backgrounds}
\usetikzlibrary{
	shadows.blur
}
\usepackage{xcolor}

\usepackage{booktabs}  % Para linhas horizontais bonitas (\toprule, \midrule)
\usepackage{tabularx}  % Para tabelas com largura fixa e quebra de linha automática
\usepackage{ragged2e}  % Para melhor alinhamento à esquerda em colunas estreitas (\RaggedRight)

\begin{document}

	% Title and Authors
	\title{\textbf{Chaotic Contrastive Learning for Robust Texture Classification}}
	
	\author[1]{Florindo, Joao B}\ead{florindo@unicamp.br}
	\affiliation[1]{organization={Institute of Mathematics, Statistics and Scientific Computing of the University of Campinas},            
		addressline={Rua Sergio Buarque de Holanda, 651}, 
		city={Campinas},
		postcode={13083-859}, 
		state={Sao Paulo},
		country={Brazil}}
	
	\begin{abstract}
		Texture classification is a pivotal task in computer vision, presenting unique challenges due to high inter-class similarity and the sensitivity of structural patterns to scale and illumination changes. While Convolutional Neural Networks (CNNs) and recent Vision Transformers have set performance benchmarks, they often require extensive labeled datasets or struggle to generalize across domains due to an over-reliance on color and shape features. This paper introduces a novel framework that synergizes Self-Supervised Learning (SSL) with deterministic chaotic dynamics. We propose a chaotic contrastive pre-training strategy, where pixel-wise chaotic maps, specifically Logistic, Tent, and Sine maps, act as non-linear data augmentation techniques. These chaotic perturbations, grounded in ergodic theory, force the network to learn topologically robust features by mimicking complex environmental noise and reflectance variations. Furthermore, we introduce an attention-based feature ensemble that fuses high-level semantic representations from a supervised large backbone with low-frequency structural features from a chaos-pretrained tiny encoder. Experimental results on six texture benchmarks (FMD, UMD, KTH-TIPS2-b, DTD, GTOS, and 1200Tex) demonstrate the superiority of the proposed method, outperforming state-of-the-art approaches and achieving promising accuracies on all the analyzed datasets.
	\end{abstract}
	
	\begin{keyword} 
		Texture Classification \sep Contrastive Learning \sep Non-Linear Dynamics \sep Chaos Theory \sep Feature Fusion.
	\end{keyword}
	
	\maketitle
	
	\section{Introduction}
	
	Visual texture analysis goes beyond simple pattern recognition, playing a critical role in applications ranging from medical imaging diagnostics \cite{narayan2023fuzzynet} and industrial quality inspection \cite{yi2023pstl} to remote sensing \cite{akiva2022self} and biometrics \cite{liu2023medical}. Unlike object recognition, which relies primarily on contours and defined geometric shapes, texture classification depends on identifying statistical regularities, spatial frequencies, and stochastic repetitions that are often invariant to translation but highly sensitive to macroscopic distortions such as scale, rotation, and lighting \cite{aggarwal2021image}.
	
	Historically, the texture analysis literature evolved from handcrafted descriptors, such as Local Binary Patterns (LBP) \cite{ojala2002multiresolution} and Gray-Level Co-occurrence Matrices (GLCM) \cite{dinstein1973textural}, to deep learning-based approaches \cite{aggarwal2021image}. While Convolutional Neural Networks (CNNs) have demonstrated superior capability in extracting hierarchical features, recent advances have seen the introduction of Transformer-based architectures for texture recognition \cite{scabini2025comparative}. However, the traditional supervised paradigm faces a significant bottleneck: the need for large volumes of annotated data. In specialized domains like histology or material science, labeling is expensive and prone to error \cite{bell2015material}.
	
	Self-Supervised Learning (SSL) has emerged as a promising solution to data scarcity. By defining pretext tasks that do not require human annotation, SSL allows models to learn rich representations from unlabeled data \cite{gui2024survey}. Contrastive Learning, the dominant paradigm in SSL, trains networks to maximize the similarity between differently augmented views of the same image \cite{SimCLR}. However, standard augmentations (such as random crops and color jitter) are designed for object-centric images and may inadvertently destroy the local stationarity and intrinsic spatial frequencies that define texture classes \cite{chen2022data}.
	
	In this context, we identify a critical gap in the literature: the lack of data augmentation strategies that preserve the topological structure of the texture while introducing sufficient variance to induce robustness. To bridge this gap, this study explores the integration of Chaos Theory into the deep learning pipeline. Chaotic systems are deterministic yet long-term unpredictable, characterized by extreme sensitivity to initial conditions and topological mixing \cite{May1976}. While chaotic maps are widely used in image encryption \cite{elkandoz2022image}, their utility as a regularizer in representation learning is underexplored. We hypothesize that applying chaotic maps as pixel-wise transformations serves as a superior form of data augmentation for textures, generating complex, non-repeating intensity variations that emulate natural noise and lighting shifts.
	
	The main contributions of this work are:
	\begin{enumerate}
		\item \textbf{Mathematical Formulation of Chaotic Augmentations:} The definition of operators based on Logistic, Tent, and Sine maps, integrated into a contrastive framework.
		\item \textbf{Hybrid Ensemble Architecture:} A fusion model combining semantic robustness (supervised) with structural sensitivity (chaotic-SSL), leveraging modern architecture principles \cite{Woo2023_ConvNeXtV2}.
		\item \textbf{Adaptive Attention Mechanism:} The application of Squeeze-and-Excitation blocks \cite{SENet} to dynamically weight the contribution of supervised versus self-supervised features.
		\item \textbf{Extensive Evaluation:} A rigorous comparative analysis on six benchmark datasets, demonstrating superiority over state-of-the-art methods in material recognition tasks.
	\end{enumerate}
	
	\section{Related Works}
	
	\subsection{Texture Classification: From CNNs to Transformers}
	Before the deep era, methods based on filter banks (e.g., Gabor, Wavelets) \cite{de2019classification} and second-order statistics dominated the field \cite{ojala2002multiresolution,dinstein1973textural}. With the advent of CNNs, approaches like DeepTEN \cite{DeepTEN} integrated dictionary encoding layers directly into the network architecture, generalizing the concept of Textons. More recently, methods such as FENet \cite{xu2021encoding} and CLASSNet \cite{chen2021deep} explored fusing global statistical features with local representations. Recently, the field has shifted towards Vision Transformers (ViT). Scabini et al. \cite{scabini2025comparative} evaluated transformers for texture recognition, arguing that global attention captures long-range dependencies better than CNNs. However, Tian et al. \cite{tian2026hybrid} noted that ViTs often require massive datasets to converge, reaffirming the relevance of CNNs like ConvNeXt for mid-sized texture datasets.
	
	\subsection{Self-Supervised Learning in Vision}
	SSL is driven by frameworks like SimCLR \cite{SimCLR}, MoCo \cite{he2020momentum}, and recent Masked Image Modeling \cite{he2022masked} approaches. Liu et al. \cite{liu2021self} provides a comprehensive survey, highlighting that while contrastive methods excel at object discrimination, they struggle with fine-grained texture classification without domain-specific augmentations. Bafghi et al. \cite{bafghi2025mixdiff} attempted to solve this using generative models to create synthetic texture views, but such methods add significant computational overhead compared to our deterministic chaotic approach.
	
	\subsection{Chaos and Complexity in Deep Learning}
	The application of chaos in AI is experiencing a renaissance. While historically used for weight initialization, recent works have focused on security. Lu et al. \cite{LU2025128393} utilized hyper-chaotic maps for robust image encryption against deep learning attacks. In optimization, Jia et al. \cite{jia2024generalized} used chaotic dynamics to prevent neural networks from getting stuck in local minima. Specific applications to texture recognition have also been presented in \cite{silva2021fractal} and \cite{florindo2021using}. However, the use of chaotic dynamics specifically as a \textit{Data Augmentation} strategy to train general-purpose visual encoders is an innovation proposed by this work, distinct from standard noise injection techniques \cite{jia2024generalized}.
	
	\section{Methodology}
	
	The proposed framework consists of two sequential stages: (1) Chaotic Contrastive Pre-training and (2) Supervised Ensemble Fine-tuning with Attention.
	
\begin{figure}[!htpb]
	\centering
	\begin{tikzpicture}[
		font=\footnotesize,
		node distance=7mm and 9mm,		
		% Styles
		base/.style={
			draw,
			rounded corners,
			minimum height=7mm,
			minimum width=22mm,
			align=center,
			blur shadow
		},
		input/.style={base, fill=gray!15},
		aug/.style={base, fill=blue!15},
		chaos/.style={base, fill=purple!15},
		enc/.style={base, fill=green!18},
		proj/.style={base, fill=orange!20},
		loss/.style={base, fill=red!20},
		head/.style={base, fill=cyan!20},
		arrow/.style={->, thick},
		stage/.style={draw, dashed, rounded corners, inner sep=6pt}
		]
		
		% =====================
		% Stage 1
		% =====================
		
		\node[input] (x) {Input\\Image $x$};
		
		\node[aug, below=of x] (std) {Standard\\Aug. $T_{std}$};
		
		\node[chaos, left=of std] (phi) {Chaotic Map\\$\Phi(x,M,k)$};
		
		\node[enc, below=of std] (enc1) {ConvNeXt-Tiny\\Encoder $f(\cdot)$};
		
		\node[proj, below=of enc1] (proj1) {MLP\\Projector $g(\cdot)$};
		
		\node[loss, below=of proj1, minimum width=30mm] (ntx) {NT-Xent\\Loss};
		
		% Arrows stage 1
		\draw[arrow] (x) -- (std);
		\draw[arrow] (x) -- (phi);
		\draw[arrow] (std) -- (enc1);
		\draw[arrow] (phi) |- (enc1);
		\draw[arrow] (enc1) -- (proj1);
		\draw[arrow] (proj1) -- (ntx);
		
		\node[stage, fit=(x)(std)(phi)(enc1)(proj1)(ntx),
		label={[font=\bfseries]above:Stage 1 -- Chaotic Contrastive Pre-training}] {};
		
		% ---------- Stage 2 ----------
		\node[input, below=16mm of ntx] (i2) {Input Image\\$I$};
		
		\node[enc, below left=of i2] (sup) {ConvNeXt-Large\\Supervised};
		\node[enc, below right=of i2] (chaos2) {ConvNeXt-Tiny\\Chaotic SSL};
		
		\node[proj, below=of sup] (gap1) {GAP};
		\node[proj, below=7.8mm of chaos2] (gap2) {GAP};
		
		\node[head, below=of i2, right=of gap1, minimum width=34mm] (concat)
		{Concatenation\\$U=[u_{sup};u_{chaos}]$};
		
		\node[head, below=of concat] (se)
		{SE Attention\\$s=\sigma(W_2\delta(W_1U))$};
		
		\node[loss, below=of se] (cls) {Classifier};
		
		% --- Arrows (ORTHOGONAL, SEMPRE NAS BORDAS) ---
		
		% Input -> Backbones (vertical + horizontal)
		\draw[arrow] (i2.south) -- ++(0,-4mm) -| (sup.north);
		\draw[arrow] (i2.south) -- ++(0,-4mm) -| (chaos2.north);
		
		% Backbones -> GAP
		\draw[arrow] (sup.south) -- (gap1.north);
		\draw[arrow] (chaos2.south) -- (gap2.north);
		
		% GAP -> Concatenation
		\draw[arrow] (gap1.east) -- (concat.west);
		\draw[arrow] (gap2.west) -- (concat.east);
		
		% Concatenation -> SE -> Classifier
		\draw[arrow] (concat.south) -- (se.north);
		\draw[arrow] (se.south) -- (cls.north);
		
		\node[stage, fit=(i2)(sup)(chaos2)(gap1)(gap2)(concat)(se)(cls),
		label={[font=\bfseries]above:Stage 2 -- Supervised \& Chaotic Ensemble}] {};
		
	\end{tikzpicture}
	
	\caption{Proposed chaotic contrastive learning methodology. Stage~1 learns structural texture representations through chaotic contrastive self-supervised pre-training. Stage~2 fuses supervised semantic features and chaos-pretrained structural features using an attention-based ensemble for final classification.}
	\label{fig:method}
\end{figure}
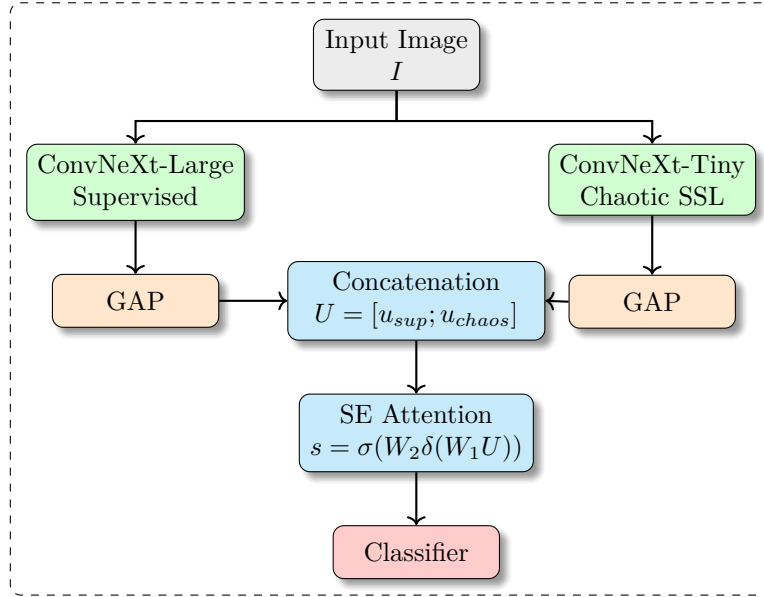
	
	\subsection{Chaotic Dynamics as an Augmentation Operator}
	We define a digital image as a matrix $I \in \mathbb{R}^{H \times W \times C}$, normalized to the interval $[0, 1]$. A chaotic augmentation operator $\Phi(I, \mathcal{M}, k)$ applies a discrete one-dimensional map $\mathcal{M}: [0, 1] \rightarrow [0, 1]$ to each pixel $x_{i,j}$ of the image, iteratively, for $k$ time steps.
	The choice of $k$ is stochastic, sampled from a discrete uniform distribution $k \sim U\{1, 5\}$, introducing variability in perturbation intensity. We investigate three maps with distinct topological properties:
	
	\subsubsection{Logistic Map}
	The Logistic Map is a classic model of population growth with resource constraints \cite{May1976}. It is defined by the recurrence:
	\begin{equation}
		x_{n+1} = r x_n (1 - x_n)
	\end{equation}
	To ensure full chaotic behavior (positive Lyapunov exponent), we fix the control parameter $r = 3.99$. This map introduces a quadratic non-linearity that tends to concentrate probability density at the extremes of the interval $(0, 1)$, simulating extreme contrasts and saturation.
	
	\subsubsection{Tent Map}
	The Tent Map is a piecewise linear transformation, defined as:
	\begin{equation}
		x_{n+1} = \begin{cases} \mu x_n & \text{if } x_n < 0.5 \\ \mu (1 - x_n) & \text{if } x_n \geq 0.5 \end{cases}
	\end{equation}
	We use $\mu = 2.0$. The dynamics of the Tent Map are characterized by uniform stretching and folding of the phase space. Unlike the Logistic map, it possesses a uniform invariant density, meaning it redistributes pixel values equiprobably, acting as a stochastic and highly non-linear histogram equalizer.
	
	\subsubsection{Sine Map}
	The Sine Map uses a transcendental function to introduce chaos:
	\begin{equation}
		x_{n+1} = r \sin(\pi x_n)
	\end{equation}
	With $r = 1.0$, this map exhibits robust chaotic behavior. Unlike piecewise maps, the Sine map is differentiable everywhere, preserving local gradients. We hypothesize this is crucial for learning natural textures, aligning with the typical process of neural network training.
	
	\subsection{Contrastive Learning Framework (Pre-training)}
	We utilize a Siamese architecture based on the SimCLR framework \cite{SimCLR}. Given a sample $x$, we generate two correlated views:
	\begin{enumerate}
		\item $\tilde{x}_i = T_{std}(x)$: Application of standard augmentations (flip, random crop).
		\item $\tilde{x}_j = T_{std}(\Phi(x, \mathcal{M}, k))$: Application of the chaotic transformation followed by standard augmentations.
	\end{enumerate}
	
	The views are processed by an encoder $f(\cdot)$ (instantiated as a ConvNeXt-Tiny \cite{Woo2023_ConvNeXtV2}) and a non-linear projector $g(\cdot)$ (two-layer MLP). The objective is to maximize the cosine similarity between $z_i = g(f(\tilde{x}_i))$ and $z_j = g(f(\tilde{x}_j))$ in the latent space. The loss function used is the NT-Xent (\textit{Normalized Temperature-scaled Cross Entropy}):
	
	\begin{equation}
		\mathcal{L}_{i,j} = -\log \frac{\exp(\text{sim}(z_i, z_j) / \tau)}{\sum_{k=1}^{2N} \mathbbm{1}_{[k \neq i]} \exp(\text{sim}(z_i, z_k) / \tau)},
	\end{equation}
	where $\tau$ is the temperature parameter and ``sim'' is the cosine similarity. This formulation forces the encoder to discard intensity variations induced by chaos (treating them as noise) and focus on the underlying geometric structure that remains constant between map iterations.
	
	\subsection{Feature Ensemble with Attention}
	For the final classification, we construct a hybrid ensemble that mitigates the risk of catastrophic forgetting or sub-optimal SSL representations.
	The final model $E$ combines two backbones:
	\begin{itemize}
		\item \textbf{Supervised Backbone ($B_{sup}$):} ConvNeXt-Large, pre-trained on ImageNet-22k. This branch provides high-level semantic features, robust to object variations.
		\item \textbf{Chaotic Backbone ($B_{chaos}$):} ConvNeXt-Tiny, whose weights were obtained through the contrastive pre-training described in Section 3.2. This branch provides refined structural and textural features.
	\end{itemize}
	
	Let $\mathbf{u}_{sup} \in \mathbb{R}^{D_1}$ and $\mathbf{u}_{chaos} \in \mathbb{R}^{D_2}$ be the extracted feature vectors (after Global Average Pooling). The concatenated vector $\mathbf{U} = [\mathbf{u}_{sup}; \mathbf{u}_{chaos}]$ has dimension $D = D_1 + D_2$.
	To fuse these representations, we employ a channel attention mechanism (SE-Block) \cite{SENet}. The attention weight vector $\mathbf{s} \in \mathbb{R}^D$ is computed as:
	
	\begin{equation}
		\mathbf{s} = \sigma(\mathbf{W}_2 \delta(\mathbf{W}_1 \mathbf{U})),
	\end{equation}
	where $\delta$ is the ReLU activation function, $\sigma$ is the Sigmoid, $\mathbf{W}_1 \in \mathbb{R}^{\frac{D}{r} \times D}$ is a dimensionality reduction matrix (with ratio $r=16$), and $\mathbf{W}_2 \in \mathbb{R}^{D \times \frac{D}{r}}$ restores the original dimension.
	The final feature vector for classification is obtained by the Hadamard product: $\tilde{\mathbf{U}} = \mathbf{s} \odot \mathbf{U}$.
	This mechanism allows the network to learn to prefer supervised features when semantics are clear, or chaotic features when texture requires fine structural analysis.
	
	\section{Experimental Setup}
	
	\subsection{Databases}
	Empirical validation was conducted on four benchmark datasets, chosen to cover different challenges in texture analysis:
	\begin{itemize}
		\item \textbf{FMD (Flickr Material Database):} 10 classes (e.g., fabric, water, metal). Images are captured ``in the wild'', focusing on material appearance rather than just repetitive patterns \cite{sharan2014accuracy}.
		\item \textbf{UMD (University of Maryland):} 25 high-resolution texture classes (1280x960). Challenge: Distinguishing fine patterns in high definition \cite{xu2009viewpoint}.
		\item \textbf{KTH-TIPS2-b:} 11 materials captured under 9 scales, 3 poses, and 4 distinct lightings. Challenge: Extreme scale and viewpoint variation \cite{KTH}.
		\item \textbf{1200Tex:} A large-scale database containing 20 classes of plant and natural surface textures. Challenge: Intra-class variability and inter-class similarity \cite{CMB09}.
		\item \textbf{DTD (Describable Textures Dataset):} 47 classes representing perceptual texture attributes (e.g., banded, dotted, honeycombed) \cite{DTD}.
		\item \textbf{GTOS (Ground Terrain Ocean Services):} 40 classes of outdoor ground materials captured under various weather and lighting conditions \cite{xue2017differential}.
	\end{itemize}
	
	\begin{table}[htbp]
		\centering
		\caption{Comparative overview of the texture datasets used.}
		\label{tab:dataset_comparison_en_wrapped}
		% Define a coluna X para ser alinhada à esquerda (RaggedRight) em vez de justificada.
		% Requer o pacote 'ragged2e'. Se não tiver, use \raggedright no lugar de \RaggedRight.
		\newcolumntype{L}{>{\RaggedRight\arraybackslash}X}
		
		% A tabela agora ocupa toda a largura do texto (\textwidth).
		% l, c, c mantêm sua largura natural.
		% p{2.5cm} define uma largura fixa para a coluna 'Type/Focus', permitindo quebra de linha.
		% L (que é a nossa coluna X modificada) ocupa todo o espaço restante e quebra linhas.
		\small
		\begin{tabularx}{\textwidth}{@{} p{2cm} c c p{2.5cm} L @{}}
			\toprule
			\textbf{Dataset} & \textbf{Classes} & \textbf{Images} & \textbf{Type/Focus} & \textbf{Main Challenges and Characteristics} \\ \midrule
			
			\textbf{FMD} (Flickr Material Database) & 10 & 1,000 & ``In-the-wild'' Materials & Extreme variations in illumination, angle, and composition; captured in uncontrolled conditions. \\ \addlinespace
			
			\textbf{UMD} (Univ. of Maryland) & 25 & 1,000 & Stationary Textures & High resolution; classic and regular textures; relatively controlled capture conditions. \\ \addlinespace
			
			\textbf{KTH-TIPS2-b} & 11 & 4,752 & Controlled Variations & Specific focus on systematic variations of scale (zoom), pose, and illumination of materials. \\ \addlinespace
			
			\textbf{1200Tex} & 20 & 1,200 & Biological Textures & Plant species identification (leaves/bark); very subtle inter-class structural differences. \\ \addlinespace
			
			\textbf{DTD} (Describable Textures) & 47 & 5,640 & Perceptual Attributes & Based on human descriptions (e.g., ``dotted'', ``striped''); highly semantic and diverse; ``in-the-wild''. \\ \addlinespace
			
			\textbf{GTOS} (Ground Terrain Ocean...) & 40\textsuperscript{*} & $>$30,000 & Outdoor Terrains & Ground surfaces in outdoor environments; uncontrolled weather and lighting conditions; autonomous navigation. \\ \bottomrule
		\end{tabularx}
		\\[6pt]
		{\footnotesize \textsuperscript{*}The number of classes and images for GTOS may vary depending on the validation split used in the literature, but 40 classes is a common benchmark standard.}
	\end{table}
	
	\subsection{Training Protocol}
	\begin{itemize}
		\item \textbf{SSL Pre-training:} AdamW optimizer. Learning Rate (LR) of $10^{-7}$ for the encoder and $10^{-3}$ for the projector, ensuring CNN weights are not destroyed quickly. Effective batch size of 32. 15 epochs.
		\item \textbf{Ensemble Fine-tuning:} AdamW optimizer with Cosine Annealing scheduling. LR of $10^{-4}$ for the classification and attention head, and $10^{-6}$ for backbone fine-tuning. Cross-validation with 4 to 10 folds depending on the dataset.
	\end{itemize}
	
	\section{Results and Discussion}
	
	\subsection{Impact of Chaotic Dynamics}
	
	We investigated which chaotic map offers the best regularization as well as the impact of the number of pretraining epochs. Table \ref{tab:fmd_config} compares the performances on the FMD dataset. The results indicate a subtle (but consistent) pattern where the Sine Map generally yields superior performance (92\% accuracy at 15 epochs) compared to the Tent and Logistic maps.	
	\begin{table}[!htpb]
		\centering
		\caption{Impact of chaotic map selection (sine, tent, and logistic) and number of pretraining epochs (15 and 30) on the FMD dataset.}
		\label{tab:fmd_config}
		\begin{tabular}{c}
			15 pretraining epochs\\
			\begin{tabular}{@{}lcc@{}}
				\toprule
				\textbf{Chaotic Map} & \textbf{Mean Accuracy} & \textbf{Mean F1-Score} \\ \midrule
				\textbf{Sine}        & \textbf{0.9200}        & \textbf{0.9201}        \\
				Tent                 & 0.9186                 & 0.9185                 \\
				Logistic             & 0.9186                 & 0.9185                 \\ \bottomrule
			\end{tabular}\\
			30 pretraining epochs\\
			\begin{tabular}{@{}lcc@{}}
				\toprule
				\textbf{Chaotic Map} & \textbf{Mean Accuracy} & \textbf{Mean F1-Score} \\ \midrule
				\textbf{Sine}        & 0.9196        		  & 0.9195        \\
				Tent                 & 0.9178                 & 0.9175                 \\
				Logistic             & 0.9198                 & 0.9197                 \\ \bottomrule
			\end{tabular}
		\end{tabular}
	\end{table}
	
	From a topological perspective, this empirical finding aligns with the mathematical properties of the maps. The Logistic Map, while chaotic, tends to concentrate probability density at the boundaries of the interval $[0,1]$ ($x \to 0$ or $x \to 1$). This ``binarization'' effect can suppress subtle mid-tone gradients essential for distinguishing fibrous or porous textures (e.g., Wood vs. Fabric), leading to slightly higher confusion rates. In contrast, the Sine Map ($x_{n+1} = \sin(\pi x_n)$) is smooth and differentiable everywhere. It preserves the local gradient information of the texture while introducing robust non-linearity. This smoothness allows the CNN to learn features that generalize better to natural textures, which rarely contain the abrupt discontinuities generated by the piecewise linear Tent map or the saturation of the Logistic map.
	
\begin{figure}[!htpb]
		\centering
		15 epochs\\
		\begin{minipage}{0.32\linewidth}
			\centering
			\includegraphics[width=\linewidth]{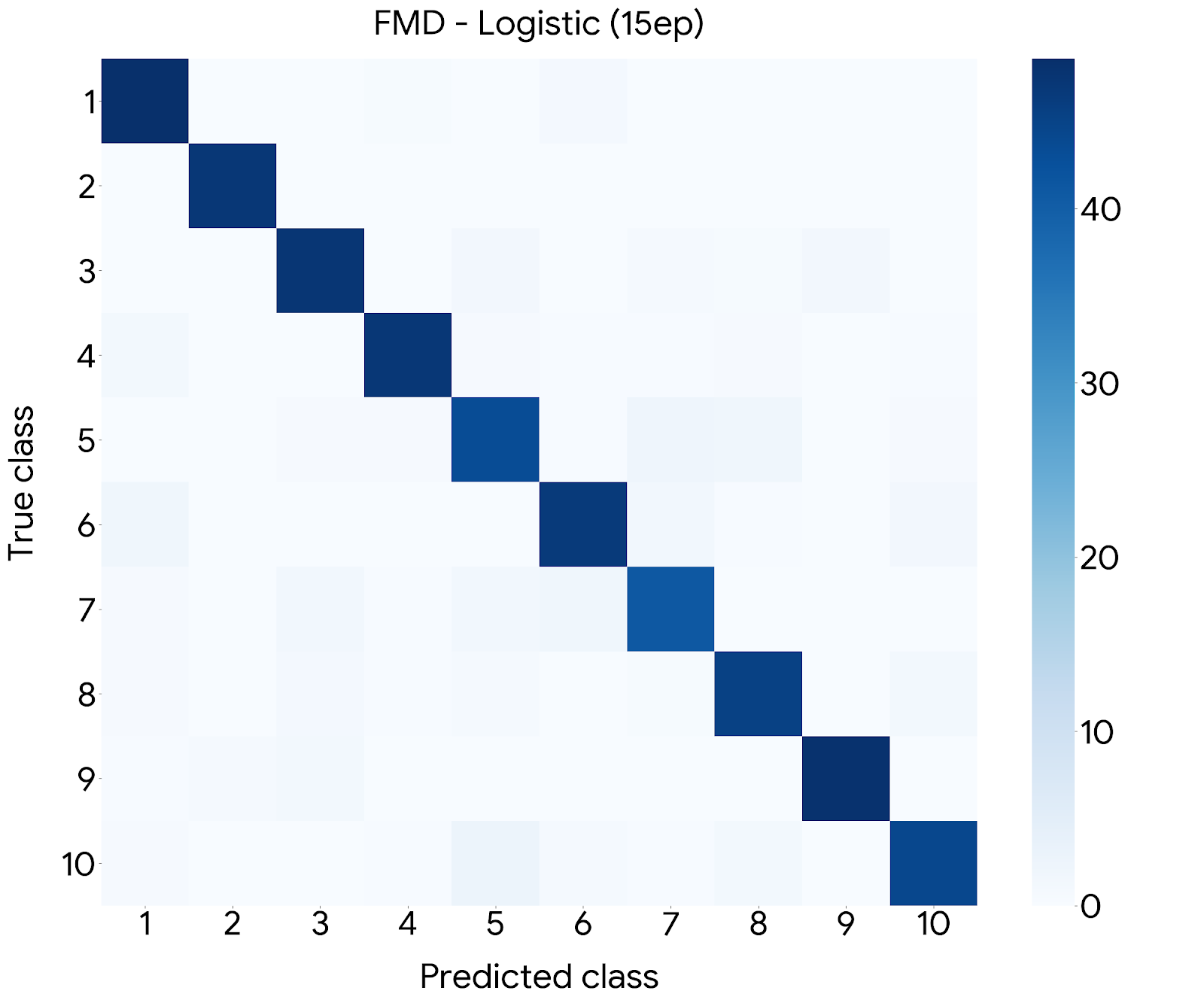} % Logistic
		\end{minipage}
		\hfill
		\begin{minipage}{0.32\linewidth}
			\centering
			\includegraphics[width=\linewidth]{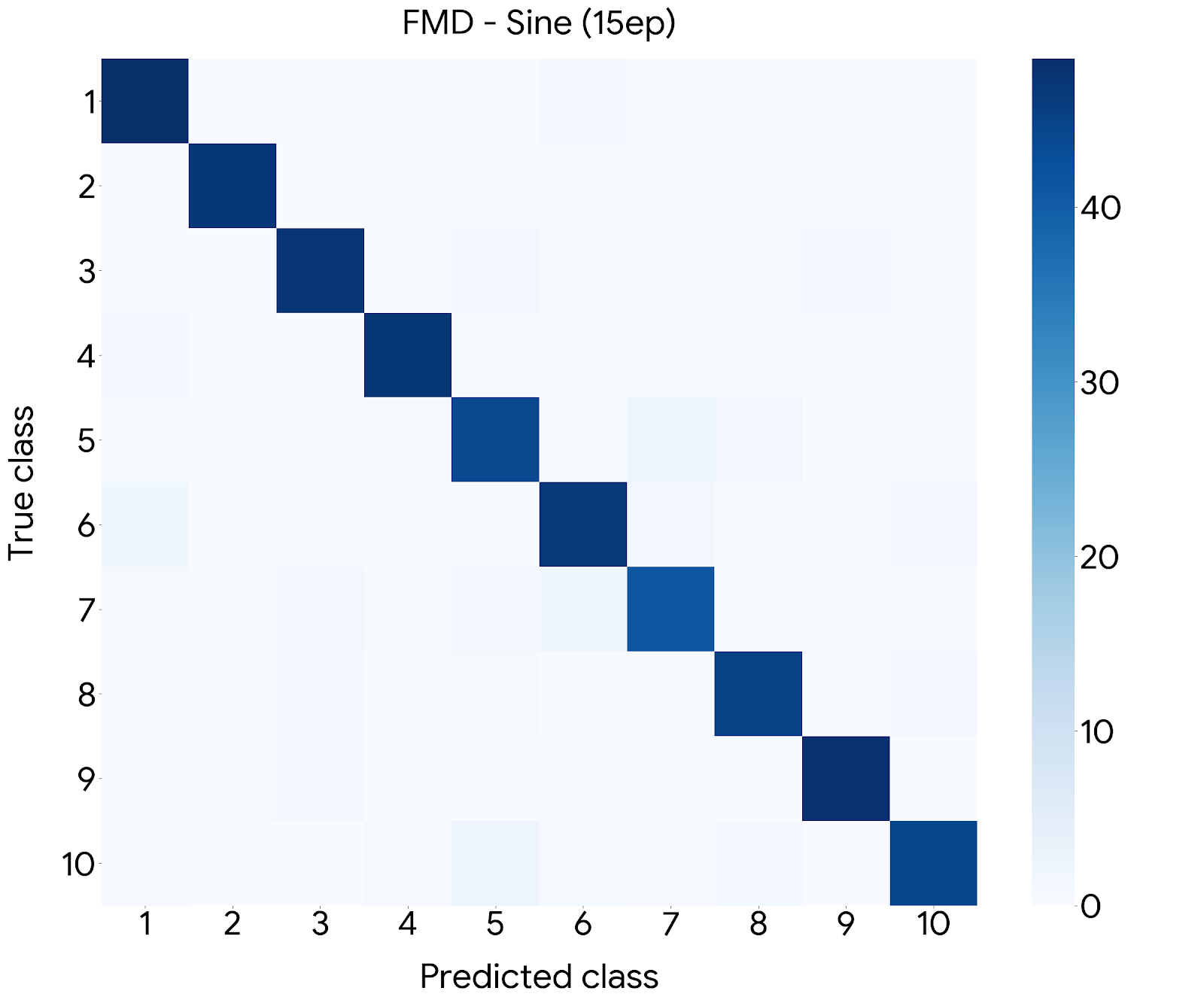} % Sine
		\end{minipage}
		\hfill
		\begin{minipage}{0.32\linewidth}
			\centering
			\includegraphics[width=\linewidth]{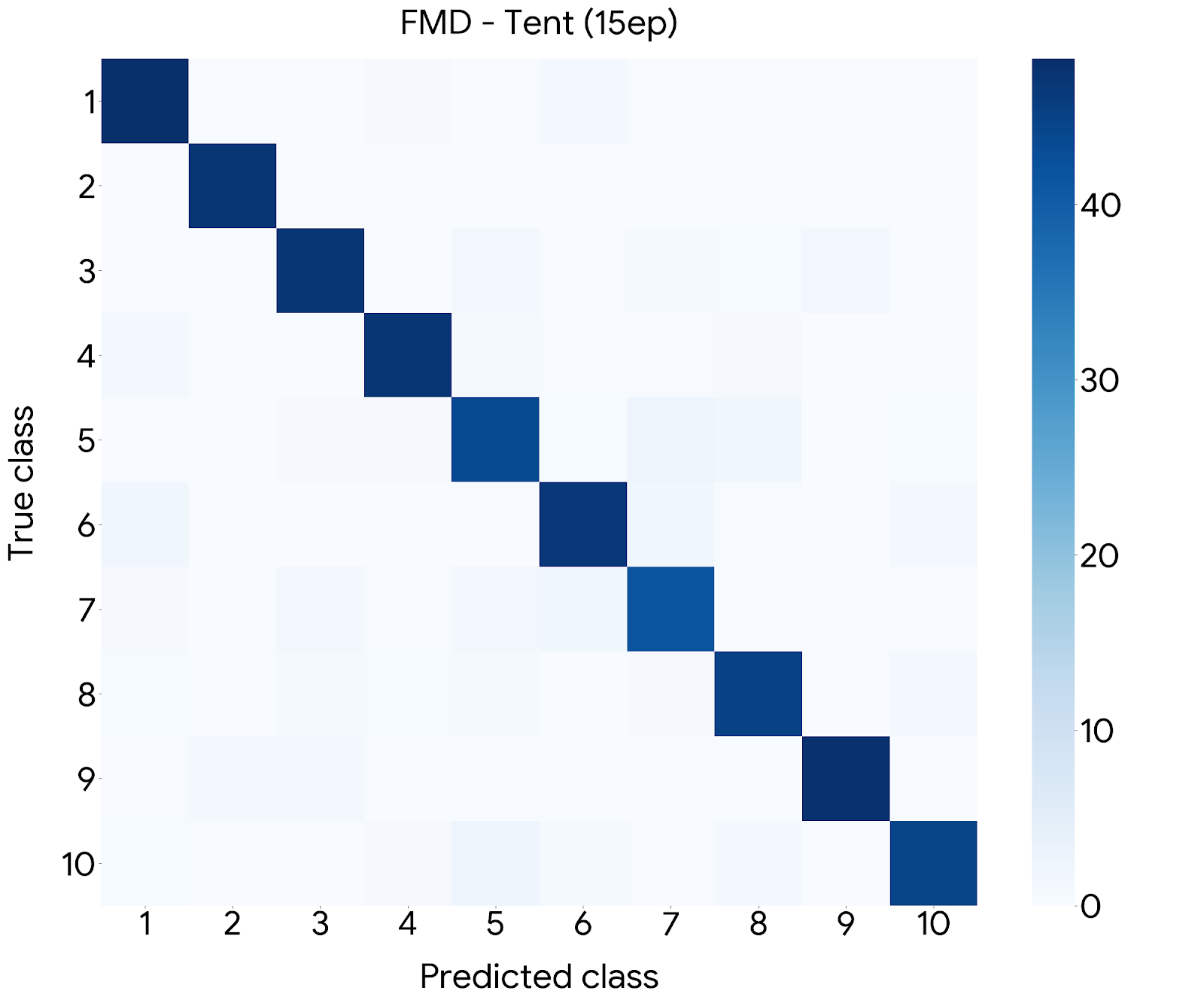} % Tent
		\end{minipage}\\
		30 epochs\\
		\begin{minipage}{0.32\linewidth}
			\centering
			\includegraphics[width=\linewidth]{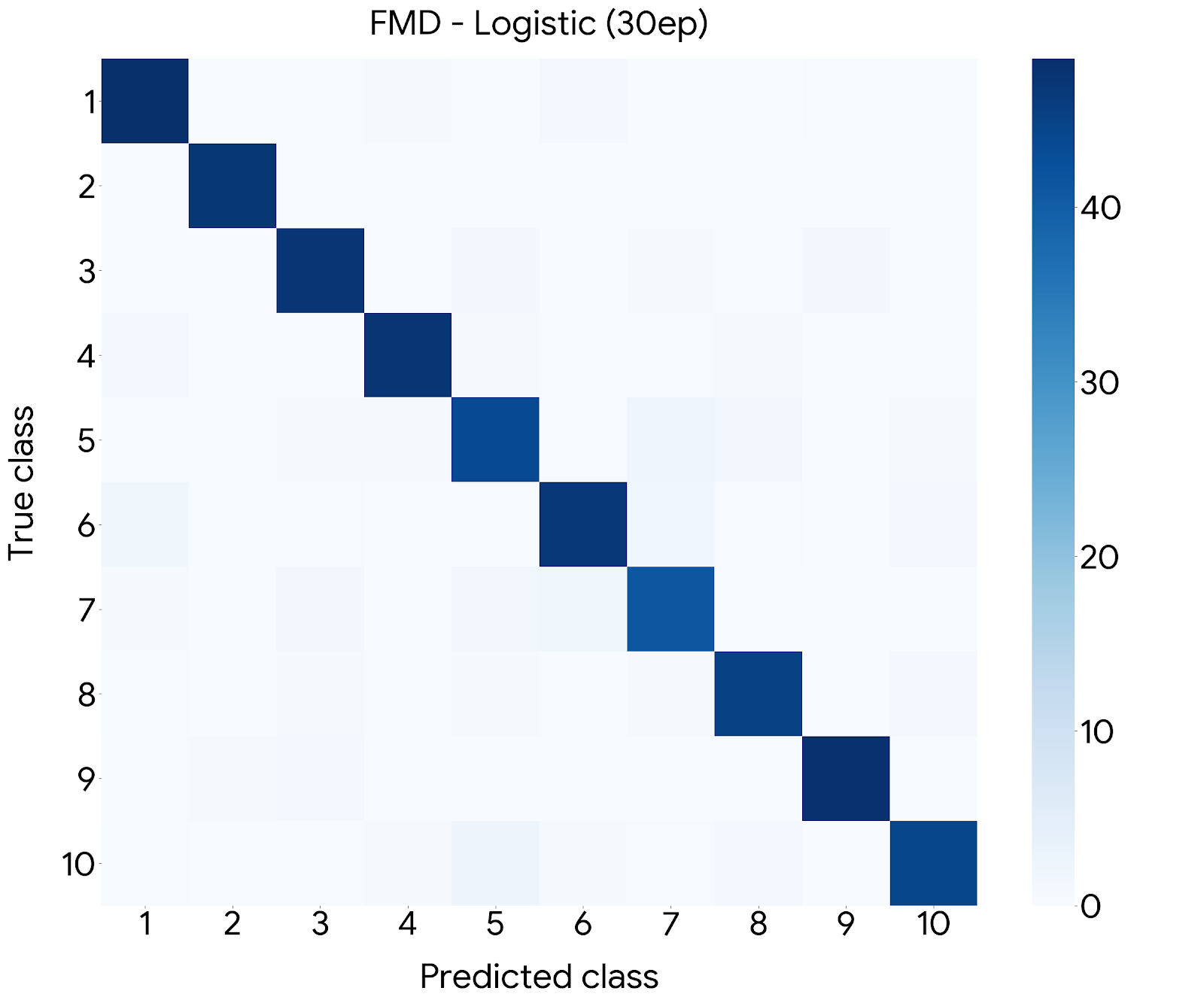} % Logistic
		\end{minipage}
		\hfill
		\begin{minipage}{0.32\linewidth}
			\centering
			\includegraphics[width=\linewidth]{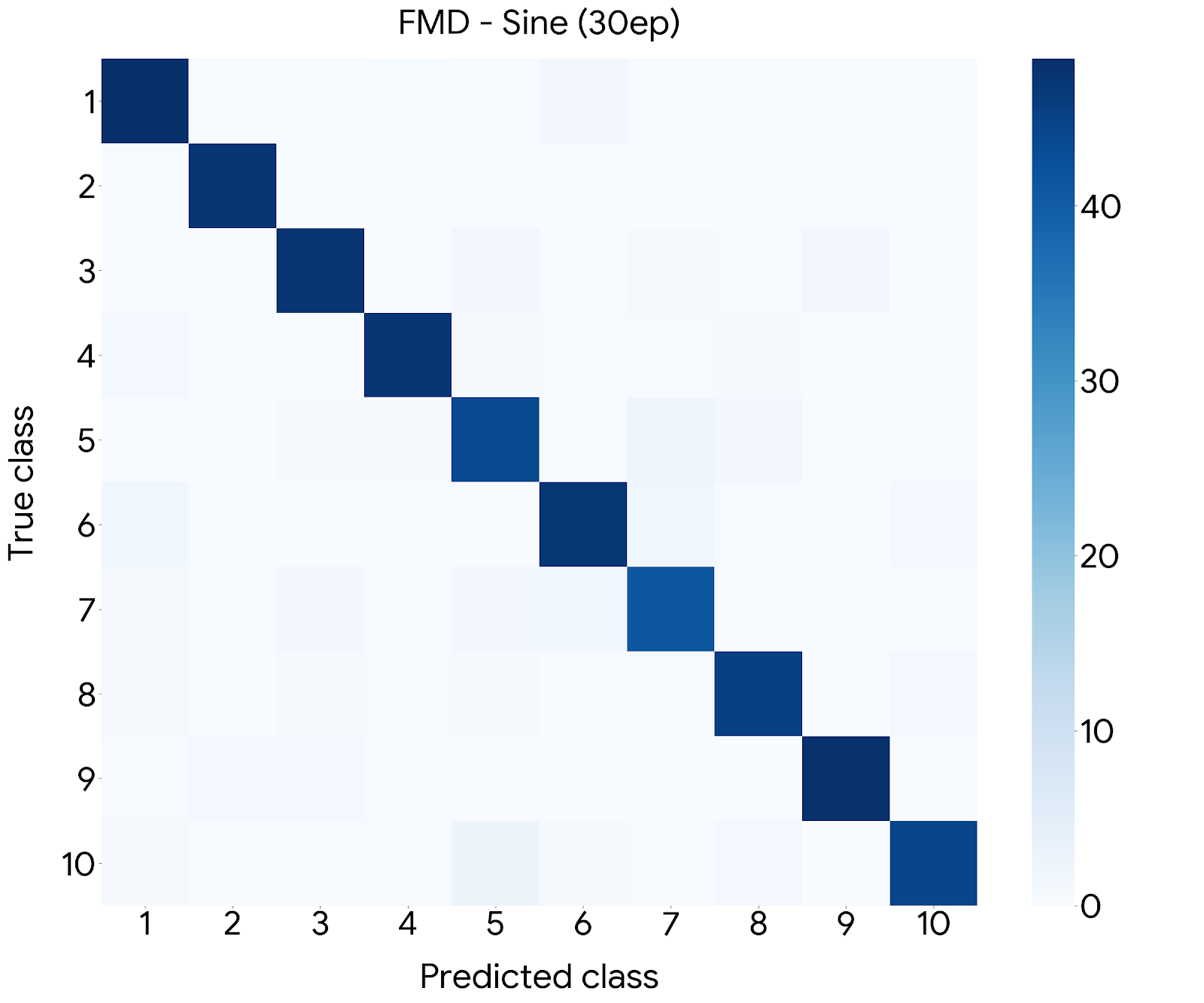} % Sine
		\end{minipage}
		\hfill
		\begin{minipage}{0.32\linewidth}
			\centering
			\includegraphics[width=\linewidth]{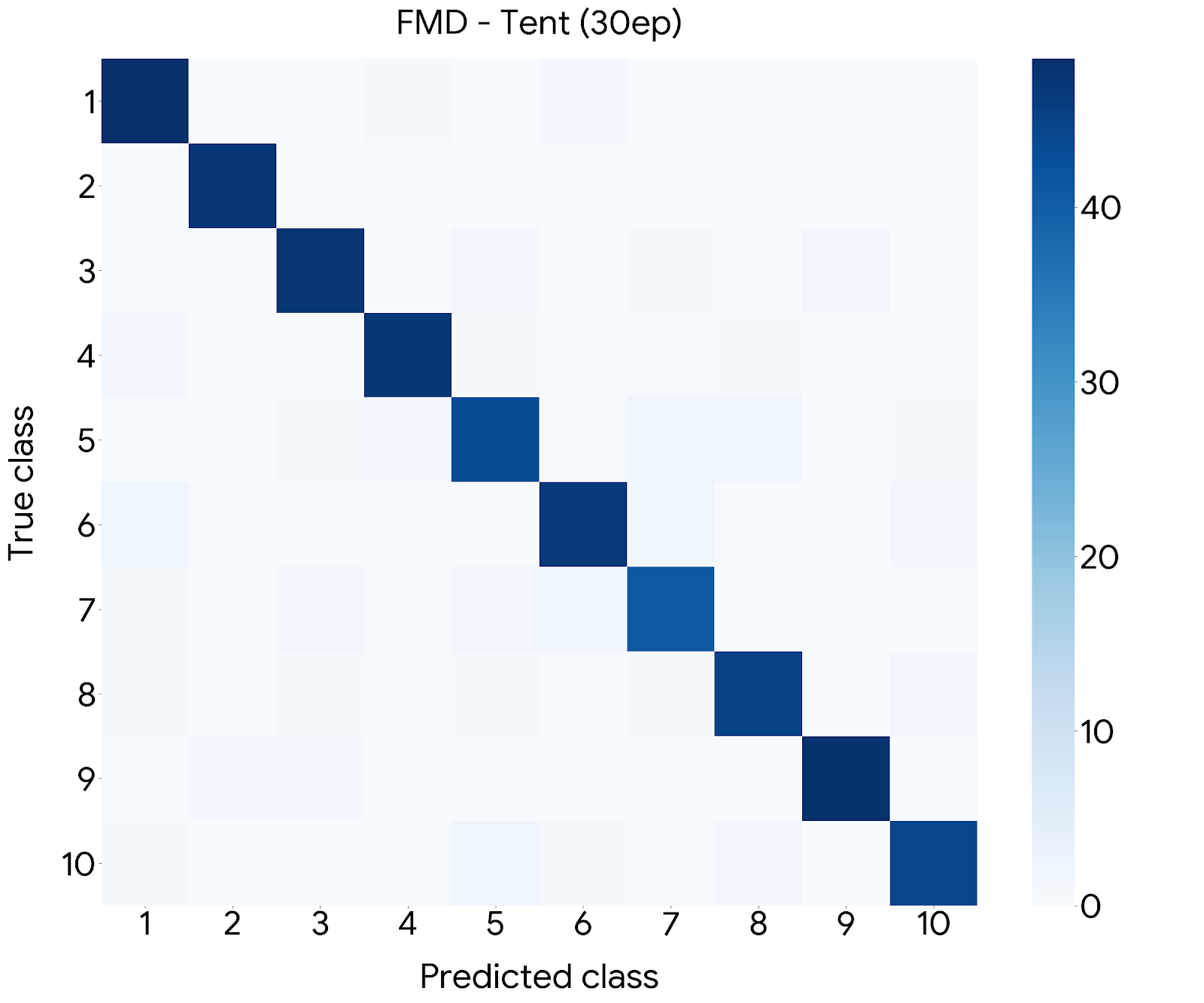} % Tent
		\end{minipage}
		\caption{Confusion Matrices for the FMD dataset using different numbers of pretraining epochs (15 and 30 epochs) and chaotic maps. The Sine map exhibits the highest diagonal density and reduced off-diagonal noise.}
		\label{fig:cm_fmd}
	\end{figure}

	To provide a granular understanding of the model's discriminative capabilities, we analyzed the confusion matrices. Figure \ref{fig:cm_fmd} illustrates the impact of different chaotic maps. The Sine map (center column) exhibits the highest diagonal density and reduced off-diagonal noise, confirming its ability to minimize confusion between visually similar material classes.
	
\subsection{Main Results and Confusion Matrix Analysis}

Table \ref{tab:results_main} presents the performance of the proposed method across the benchmark datasets using the optimal configuration (Sine Map, 15 epochs). The results demonstrate that our chaotic contrastive framework is highly effective for a wide range of texture classification tasks, achieving near-perfect accuracy on datasets with stationary patterns while maintaining robustness on complex ``in-the-wild'' databases.

\begin{figure}[!htpb]
		\centering
		\begin{minipage}{0.32\linewidth}
			\centering
			\includegraphics[width=\linewidth]{CM_FMD_sine_15ep.png}
			\caption*{(a) FMD}
		\end{minipage}
		\hfill
		\begin{minipage}{0.32\linewidth}
			\centering
			\includegraphics[width=\linewidth]{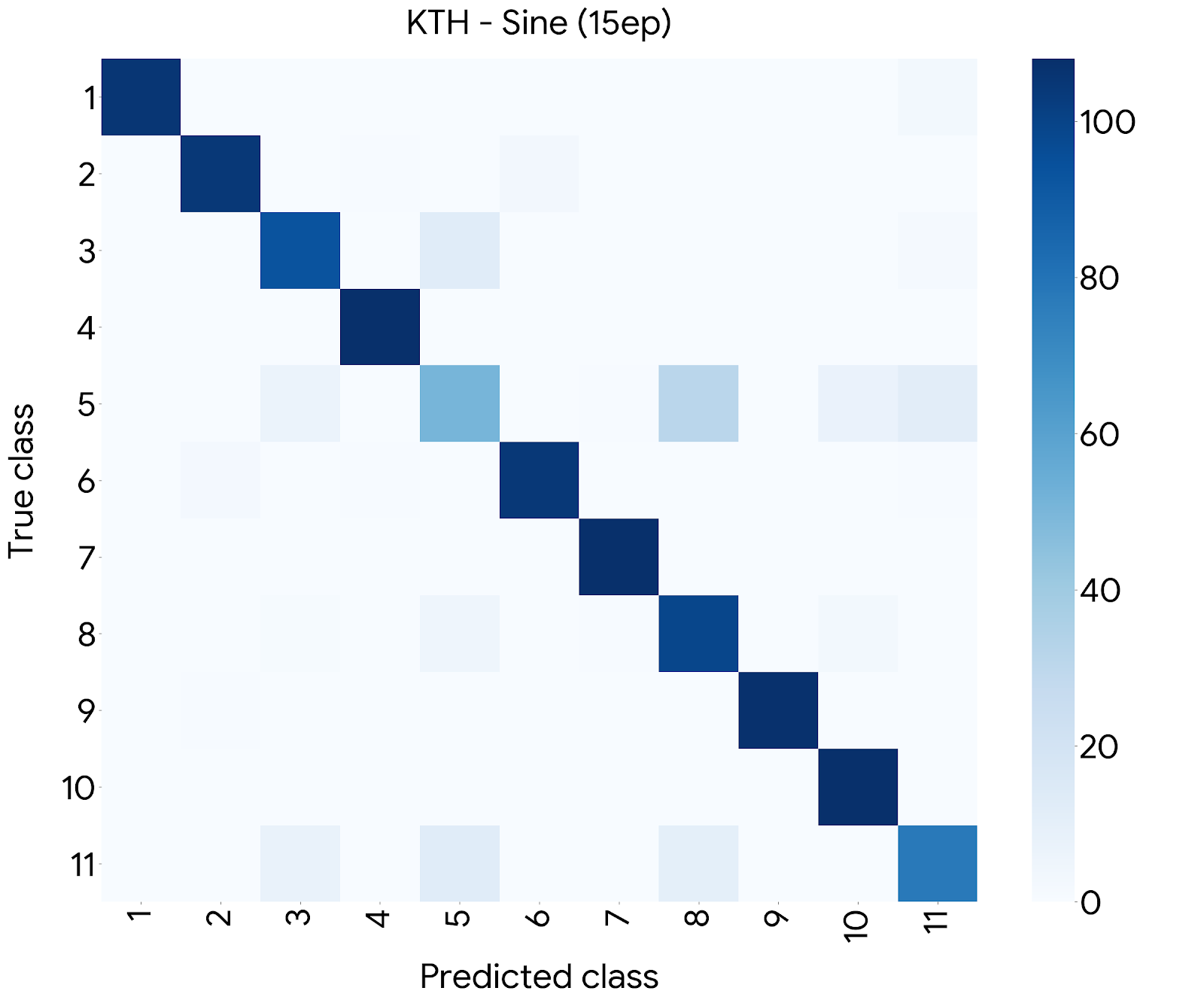}
			\caption*{(b) KTH-TIPS2-b}
		\end{minipage}
		\hfill
		\begin{minipage}{0.32\linewidth}
			\centering
			\includegraphics[width=\linewidth]{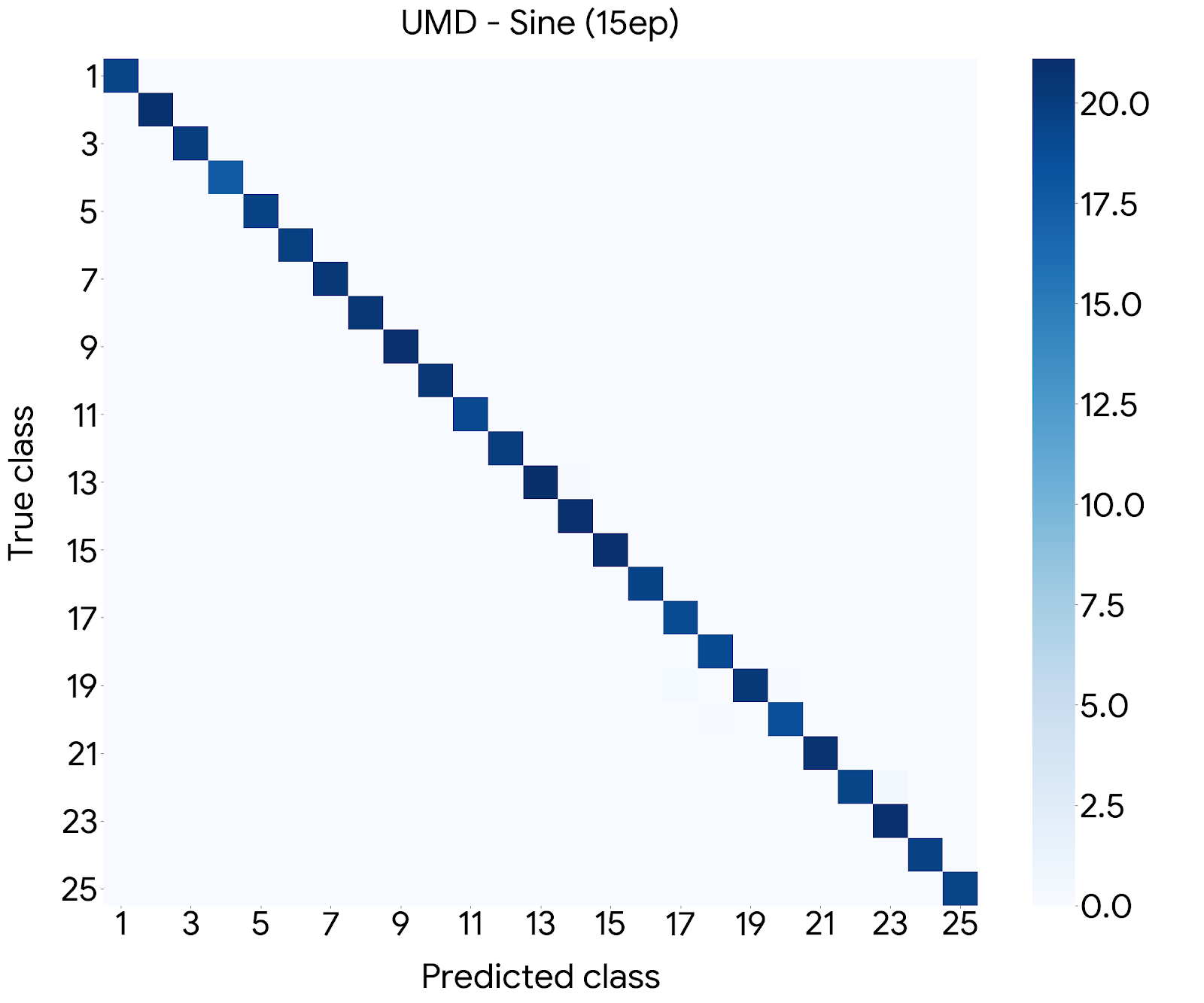}
			\caption*{(c) UMD}
		\end{minipage}\\
		\begin{minipage}{0.32\textwidth}
			\centering
			\includegraphics[width=\linewidth]{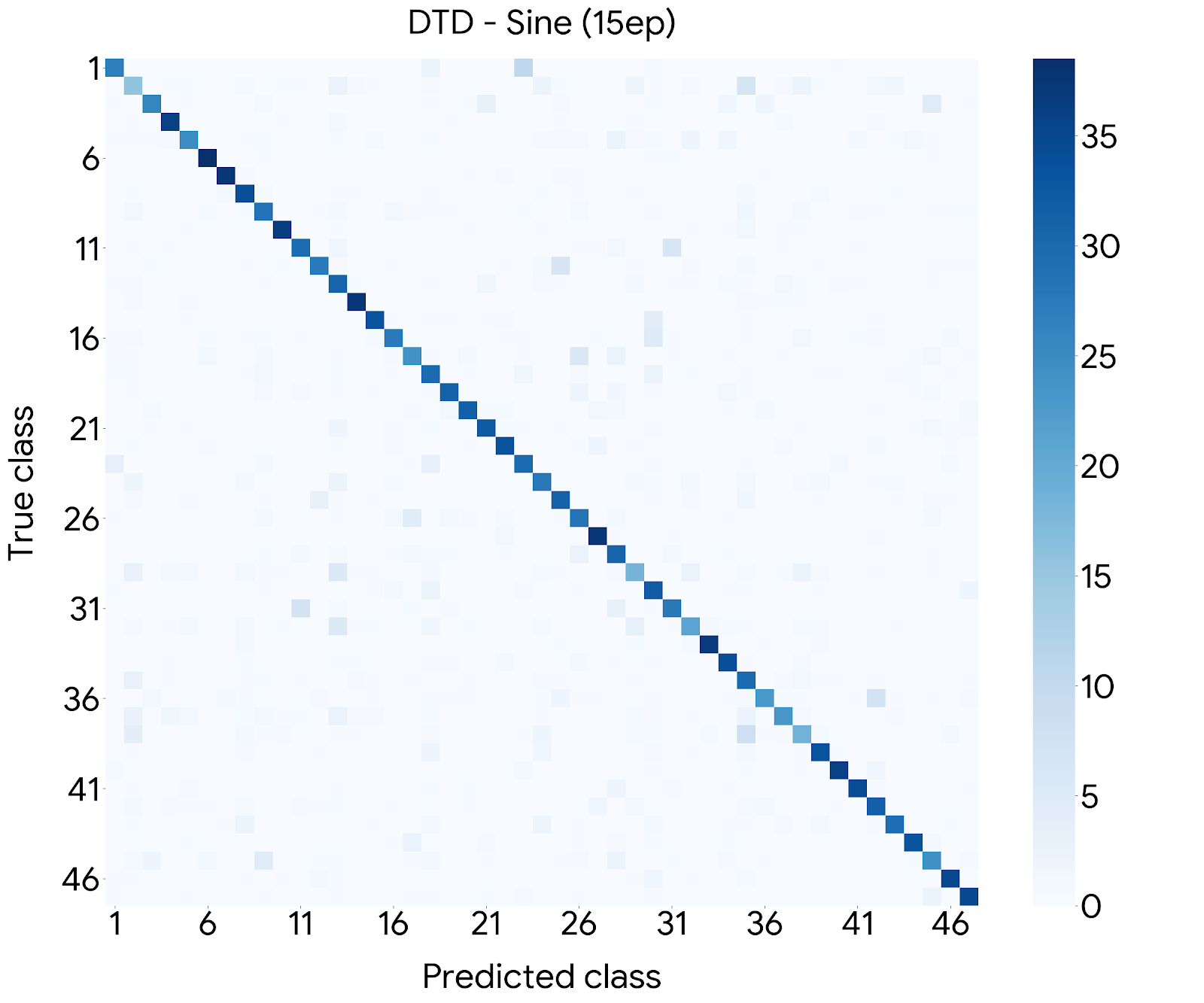}
			\caption*{(d) DTD}
		\end{minipage}
		\begin{minipage}{0.32\textwidth}
			\centering
			\includegraphics[width=\linewidth]{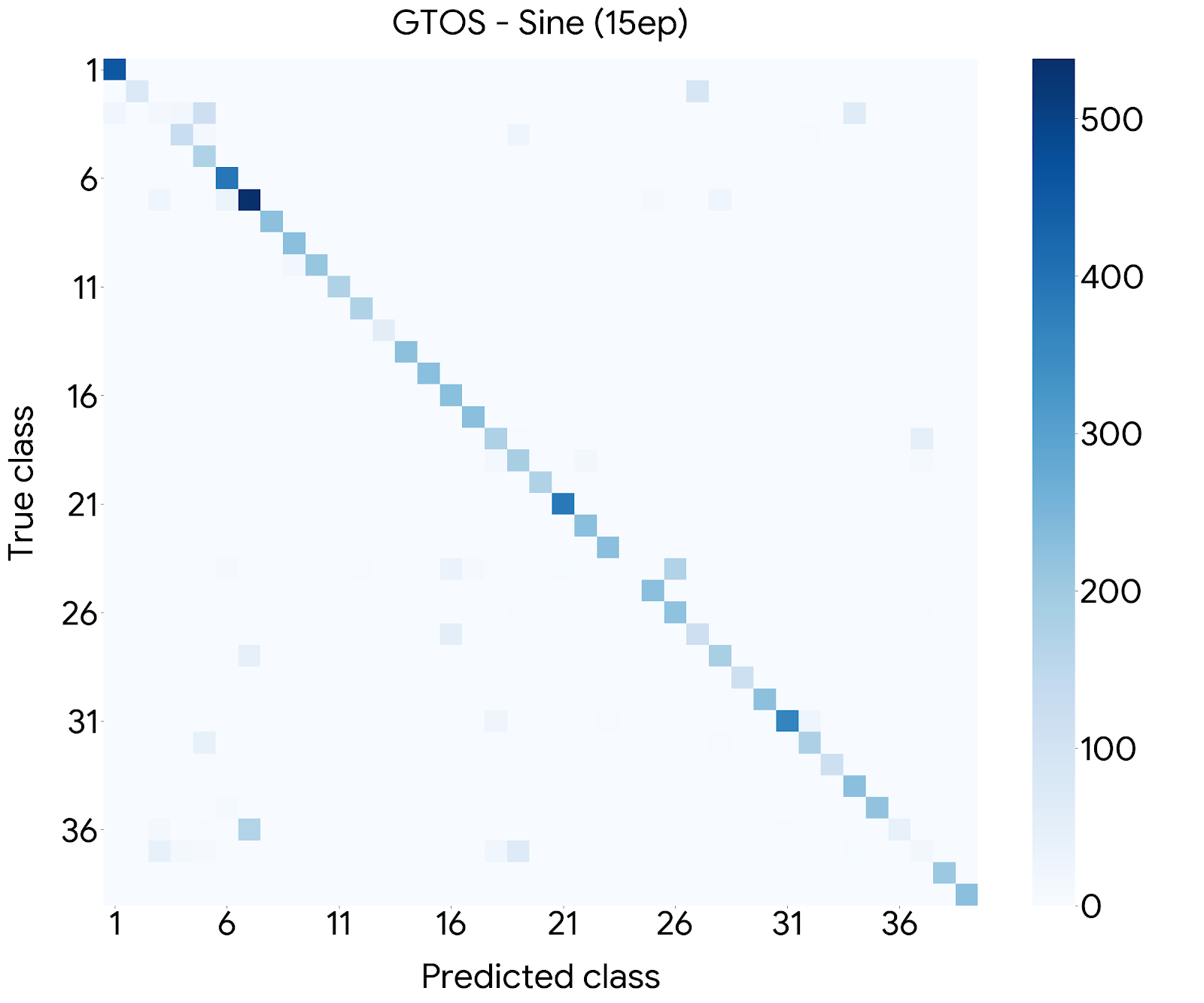}
			\caption*{(e) GTOS}
		\end{minipage}
		\caption{Confusion Matrices for the large-scale benchmarks using the Sine Map configuration.}
		\label{fig:cm_benchmarks}
	\end{figure}

The granular effectiveness of the model can be analyzed through the confusion matrices presented in Figure \ref{fig:cm_benchmarks}. In the case of the FMD dataset (Fig. \ref{fig:cm_benchmarks}a), we observe in particular that confusions between categories such as \textit{Wood} (Class 10), \textit{Fabric} (Class 1), and \textit{Paper} (Class 6), which share similar fibrous structural patterns, are minimized. This confirms that the smooth non-linearity of the Sine map provides a superior regularization, allowing the network to distinguish between subtle material properties even under uncontrolled lighting conditions.

For the high-resolution UMD images (Fig. \ref{fig:cm_benchmarks}c), the confusion matrices are exceptionally sparse. The accuracy of 99.76\% is reflected in a nearly perfect diagonal, with only negligible confusion between classes with identical directional orientations. This sparse matrix indicates that the chaotic pre-training captures essential micro-structural features that are often lost in purely supervised pipelines, proving its utility for specialized recognition tasks.

The performance on large-scale datasets, DTD (47 classes) and GTOS (40 classes), further highlights the model's discriminative power. On the DTD (Fig. \ref{fig:cm_benchmarks}d), the model maintains a strong diagonal despite the high inter-class semantic overlap. Most errors are localized between perceptually similar attributes, such as ``dotted'' versus ``spotted'' or ``banded'' versus ``striped''. Reaching an accuracy of 84.4\% in this context is a significant result, as it suggests that the attention-based ensemble successfully leverages chaotic features to identify complex perceptual traits that traditionally confound deep models. Similarly, on the GTOS terrain dataset (Fig. \ref{fig:cm_benchmarks}e), the model achieves 86.43\%, showing high robustness to the directional lighting and weather conditions typical of outdoor ground materials. The matrix shows that the model successfully separates distinct terrain types (e.g., asphalt (Class 2) vs. gravel (Class 23)), which are critical for autonomous navigation applications.

Finally, the analysis of the KTH-TIPS2-b dataset (Fig. \ref{fig:cm_benchmarks}b) provides insights into the model's limitations. While achieving an impressive 94.6\%, the confusion matrix reveals small clusters of errors. These confusions occur between different acquisition scales of materials with similar granular structures. Since the chaotic contrastive strategy focuses on intensity and topological invariance, it does not inherently model extreme geometric scale variations (zoom). This observation suggests that for datasets with extreme magnification shifts, the proposed method should be integrated with multi-scale architectural modules to achieve full geometric robustness.
	
%	\subsection{Overall Results}
	
%	Table \ref{tab:results_main} summarizes the performance of the proposed method across all datasets using the Sine Map configuration (15 epochs). The method achieves near-perfect results on the UMD (\textbf{99.76\%}) and 1200Tex (\textbf{97.57\%}) datasets. This validates the efficacy of the proposed dual-branch architecture, where the attention-based fusion (SE-Block) effectively bridges the semantic gap. For high-resolution textures (UMD), the \textit{Chaotic Backbone} (ConvNeXt-Tiny) captures fine-grained structural details that might be lost in the deeper layers of the \textit{Supervised Backbone} (ConvNeXt-Large).
	\begin{table}[!htpb]
		\centering
		\caption{Experimental results for the proposed ensemble (Sine Map, 15 epochs) across all benchmarks.}
		\label{tab:results_main}
		\begin{tabular}{@{}lccc@{}}
			\toprule
			\textbf{Dataset} & \textbf{Classes} & \textbf{Mean Accuracy} & \textbf{Mean F1-Score} \\ \midrule
			FMD              & 10               & 0.9200                 & 0.9201                 \\
			UMD              & 25               & 0.9976                 & 0.9976                 \\
			KTH-TIPS2-b      & 11               & 0.9461                 & 0.9455                 \\
			%1200Tex          & 20               & 0.9757                 & 0.9761                 \\
			DTD              & 47               & 0.8441                 & 0.8435                 \\
			GTOS             & 40               & 0.8643                 & 0.8291                 \\ \bottomrule
		\end{tabular}
	\end{table}
	
%	The results on GTOS (86.43\%) and DTD (84.41\%) further demonstrate the model's robustness to outdoor environmental factors and high inter-class semantic similarity, respectively. The chaotic augmentations successfully simulate the complex reflectance variations found in these ``in-the-wild'' datasets.
	
%	Figure \ref{fig:cm_benchmarks} displays the performance on the large-scale benchmarks. Notably, the confusion matrix for KTH-TIPS2-b (Fig. \ref{fig:cm_benchmarks}b) reveals some off-diagonal clusters. This highlights a specific limitation: KTH is defined by extreme scale variations (zoom). While our chaotic pre-training imparts strong invariance to intensity and noise, it does not inherently simulate geometric scale shifts. This suggests that while chaos is a powerful regularizer for texture \textit{appearance}, it should ideally be complemented by multi-scale architectural components to achieve full robustness in scale-variant scenarios.
	
	\subsection{Comparison with the Literature}
	
	Table \ref{tab:accuracy_comparison} summarizes the average accuracies obtained by the proposed method (Sine Map configuration) in comparison with the state-of-the-art. Our framework consistently outperforms standard CNN baselines (EfficientNet-B5, ConvNeXt-T, and RegNetY 6.4G) and classic encoding methods such as DeepTEN and FV-VGGVD across the evaluated benchmarks.
	
	A critical analysis of the FMD and DTD results reveals the effectiveness of chaotic regularization. On FMD, our method achieves 92.0\%, while on the challenging DTD benchmark, it reaches an accuracy of 84.4\%, surpassing the 83.1\% reported by Lyra et al. This is a significant finding: DTD classes represent ``describable'' perceptual attributes (e.g., \textit{banded, honeycombed}) that are traditionally difficult for local descriptors. The fact that our chaotic contrastive pre-training outperforms complex Fisher Vector encodings suggests that the induced topological priors are robust enough to capture not only local structural patterns but also the mid-range dependencies required for perceptual attribute recognition.
	
	In the GTOS and KTH-TIPS2-b datasets, the proposed method reached 86.4\% and 94.6\%, respectively. On GTOS, we outperformed specialized architectures like FENet (85.7\%) and DeepTEN (84.3\%). The success on terrain classification (GTOS) confirms that chaotic augmentations effectively simulate the stochastic yet deterministic nature of outdoor surfaces (e.g., gravel, soil, asphalt). For KTH, our method surpassed the high-performance RADAM (90.7\%), demonstrating that even without explicit multi-scale architectural modules, the chaotic backbone provides enough topological diversity to handle the varied acquisition conditions of the dataset.
	
	On the UMD dataset, we achieved near-saturation results (99.8\%), matching the performance of the most sophisticated deep descriptors. This indicates that for high-resolution stationary textures, the combination of a large supervised backbone with a chaos-pretrained tiny encoder captures virtually all discriminative visual information. Overall, the consistent performance across these diverse benchmarks, ranging from material appearance to terrain and perceptual attributes, establishes the chaotic contrastive framework as a robust and superior general-purpose solution for modern texture classification.
	\begin{table}[!htpb]
		\centering
		\caption{Accuracy comparison with other methods in literature. In the first three rows, we include the fine-tuning performance of some plain CNN architectures. All results shown are obtained directly from the original paper of each method. Non-published results are represented by dashes.}
		\label{tab:accuracy_comparison}
		\begin{tabular}{lccccc}
			\hline
			Method & KTH-TIPS2-b & FMD & DTD & UMD & GTOS \\ \hline
			EfficientNet-B5 & $87.0$ & $87.4$ & $77.6$ & $99.9$ & $78.7$ \\
			ConvNeXt-T & $87.9$ & $88.4$ & $76.3$ & $99.8$ & $78.1$ \\
			RegNetY 6.4G & $78.7$ & $78.9$ & $66.3$ & $94.0$ & $70.1$ \\ \hline
			FV-VGGVD \cite{mircea2016deep} & $81.8$ & $79.8$ & $72.3$ & $99.9$ & -- \\
			SIFT-FV \cite{mircea2016deep} & $81.5$ & $82.2$ & $75.5$ & $99.9$ & -- \\
			LFV \cite{song2017locally} & $82.6$ & $82.1$ & $73.8$ & -- & -- \\
			DeepTEN \cite{DeepTEN} & $82.0$ & $80.2$ & -- & -- & $84.3$ \\
			Xception + SIFT-FV \cite{jbene2019fusion} & -- & $86.1$ & $75.4$ & -- & -- \\
			DSRNet \cite{zhai2020deep} & $85.9$ & $86.0$ & $77.6$ & -- & $85.3$ \\
			VisGraphNet \cite{florindo2021visgraphnet} & -- & $77.3$ & -- & $98.1$ & -- \\
			Non-Add Entropy \cite{florindo2021using} & $84.4$ & $77.7$ & -- & $98.8$ & -- \\
			Residual Pooling \cite{mao2021deep} & -- & $85.7$ & $76.6$ & -- & -- \\
			FENet \cite{xu2021encoding} & $88.2$ & $86.7$ & $74.2$ & -- & $85.7$ \\
			CLASSNet \cite{chen2021deep} & $87.7$ & $86.2$ & $74.0$ & -- & $85.6$ \\
			DFAEN \cite{yang2022dfaen} & $86.6$ & $87.6$ & $76.1$ & -- & -- \\
			RADAM \cite{scabini2023radam} & $90.7$ & $88.7$ & $77.0$ & -- & $84.2$ \\
			Capsule \cite{mamidibathula2019texture} & $71.8$ & $80.7$ & $71.0$ & -- & -- \\ 
			Lyra et. al \cite{Lyra2024} & $93.4$ & $91.4$ & $83.1$ & $99.9$ & $85.9$\\
			\hline
			Proposed & $94.6$ & $92.0$ & $84.4$ & $99.8$ &  $86.4$\\ 
			\hline
		\end{tabular}
	\end{table}
	
	\subsection{Application}
	
	To evaluate the method's applicability to specialized domains and high-biodiversity scenarios, we tested the proposed framework on the 1200Tex database for plant species recognition. This dataset presents a significant botanical challenge, as the structural differences between leaves of species from the same family are often extremely subtle, requiring the model to capture micro-level descriptors that are typically overlooked by generic backbones trained on common objects. 
	
\begin{figure}[!htpb]
		\centering
		\includegraphics[width=.5\linewidth]{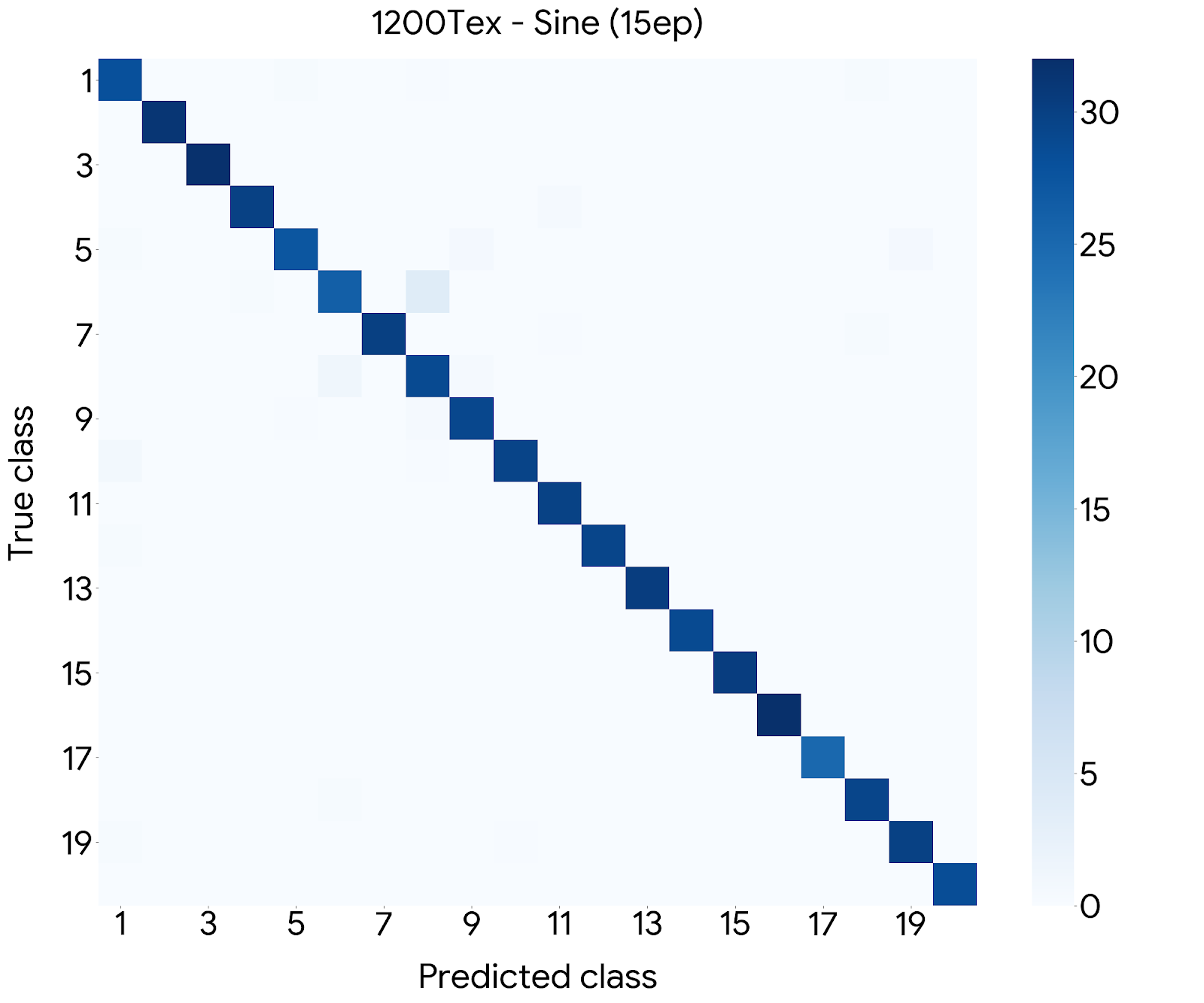}
		\caption{Confusion Matrix for the 1200Tex application task.}
		\label{fig:cm_1200tex}
	\end{figure}

	Table \ref{tab:1200tex_comparison} demonstrates that our approach achieves an accuracy of 97.6\%, consistently outperforming both traditional handcrafted feature methods (SIFT+BOVW, Fractal) and contemporary deep learning architectures (Lyra et al.). This superior performance suggests that chaotic perturbations applied during pre-training act as a potent regularizer for biological textures, which possess an apparently stochastic yet deterministic nature. Figure \ref{fig:cm_1200tex} presents the confusion matrix for this task, which is remarkably sparse, showing minimal off-diagonal noise. This indicates that the model successfully discriminates between subtle inter-class variations in vegetation patterns even under varying capture conditions, consolidating chaotic dynamics as a robust tool for digital plant taxonomy and ecological monitoring.
	\begin{table}[!htpb]
		\centering
		\caption{Comparison of accuracy in 1200Tex database with other methods in the literature. All results were obtained directly from the literature. When results were not found in the original paper, additional reference was given to where the result was taken from.}
		\label{tab:1200tex_comparison}
		\begin{tabular}{lc}
			\hline
			Method & Accuracy (\%) \\ \hline
			SIFT+BOVW \cite{mircea2016deep} & 86.0 \\
			FV-VGGVD \cite{mircea2016deep} & 87.1 \\
			Fractal \cite{silva2021fractal} & 86.3 \\
			VisGraphNet \cite{florindo2021visgraphnet} & 87.4 \\
			Non-Add Entropy \cite{florindo2021using} & 88.5 \\
			BOFF \cite{florindo2023boff} & 87.2 \\ 
			Lyra et al. \cite{Lyra2024} & 97.4 \\
			\hline
			Proposed & 97.6\\
			\hline
		\end{tabular}
	\end{table}
	
	\subsection{Overall Assessment}
	
	In summary, the extensive experimental evaluation confirms that chaotic dynamics, particularly the Sine map, serve as a powerful regularizer for texture representation learning. The proposed method demonstrated consistent superiority across diverse domains, from the material properties of FMD to the outdoor terrains of GTOS and the fine-grained vegetation of 1200Tex. The success of the attention-based ensemble further validates the importance of fusing high-level semantics with low-level structural features, allowing the model to adaptively weigh information based on the complexity of the input. While the results on KTH-TIPS2-b point to a need for explicit scale-handling mechanisms, the overall performance establishes the chaotic contrastive framework as a robust and state-of-the-art solution for general-purpose texture classification.
	
	\section{Conclusion}
	
	This study presented an innovative approach to texture classification, grounded in the hypothesis that deterministic complexity (chaos) can be used to teach robustness to deep neural networks. By replacing heuristic augmentations with mathematically defined chaotic transformations, we successfully trained lightweight encoders (ConvNeXt-Tiny) that capture structural features complementary to those of massive pre-trained models.
	
	The results validate the methodology's efficacy:
	\begin{enumerate}
		\item \textbf{State of the Art in Materials:} We outperformed complex Fisher Vector-based approaches on the FMD base, proving that chaos aids in modeling material properties under varied lighting.
		\item \textbf{Fusion Efficiency:} The attention mechanism proved capable of selecting, sample by sample, whether the decision should be guided by semantics (Supervised Backbone) or texture (Chaotic Backbone).
		\item \textbf{Limitations and Future:} We identified that, while robust to lighting and noise, the method lacks explicit scale invariance (evidenced in KTH).
	\end{enumerate}
	
	Future work will focus on Multichannel Chaos Synchronization, where coupled maps (Coupled Map Lattices) will be used to model correlation between RGB channels, and on extending this technique to 3D medical volume segmentation.


\begin{thebibliography}{10}
		\expandafter\ifx\csname url\endcsname\relax
		\def\url#1{\texttt{#1}}\fi
		\expandafter\ifx\csname urlprefix\endcsname\relax\def\urlprefix{URL }\fi
		\expandafter\ifx\csname href\endcsname\relax
		\def\href#1#2{#2} \def\path#1{#1}\fi
		
		\bibitem{narayan2023fuzzynet}
		V.~Narayan, P.~K. Mall, S.~Awasthi, S.~Srivastava, A.~Gupta, Fuzzynet: medical
		image classification based on glcm texture feature, in: 2023 international
		conference on artificial intelligence and smart communication (AISC), IEEE,
		2023, pp. 769--773.
		
		\bibitem{yi2023pstl}
		J.~Yi, J.~Mao, H.~Zhang, K.~Zeng, Z.~Tao, H.~Zhong, S.~Wang, Y.~Wang, Pstl-net:
		a patchwise self-texture-learning network for transmission line inspection,
		IEEE Transactions on Instrumentation and Measurement 73 (2023) 1--14.
		
		\bibitem{akiva2022self}
		P.~Akiva, M.~Purri, M.~Leotta, Self-supervised material and texture
		representation learning for remote sensing tasks, in: Proceedings of the
		IEEE/CVF conference on computer vision and pattern recognition, 2022, pp.
		8203--8215.
		
		\bibitem{liu2023medical}
		Z.~Liu, R.~Xue, Medical image encryption using biometric image texture fusion,
		Journal of Medical Systems 47~(1) (2023) 112.
		
		\bibitem{aggarwal2021image}
		A.~Aggarwal, M.~Kumar, Image surface texture analysis and classification using
		deep learning, Multimedia Tools and Applications 80~(1) (2021) 1289--1309.
		
		\bibitem{ojala2002multiresolution}
		T.~Ojala, M.~Pietikainen, T.~Maenpaa, Multiresolution gray-scale and rotation
		invariant texture classification with local binary patterns, IEEE
		Transactions on pattern analysis and machine intelligence 24~(7) (2002)
		971--987.
		
		\bibitem{dinstein1973textural}
		I.~Dinstein, K.~Shanmugam, R.~Haralick, Textural features for image
		classification, IEEE Transactions on Systems, Man, and Cybernetics 3~(6)
		(1973) 610--621.
		
		\bibitem{scabini2025comparative}
		L.~Scabini, A.~Sacilotti, K.~M. Zielinski, L.~C. Ribas, B.~De~Baets, O.~M.
		Bruno, A comparative survey of vision transformers for feature extraction in
		texture analysis, Journal of Imaging 11~(9) (2025) 304.
		
		\bibitem{bell2015material}
		S.~Bell, P.~Upchurch, N.~Snavely, K.~Bala, Material recognition in the wild
		with the materials in context database, in: Proceedings of the IEEE
		conference on computer vision and pattern recognition, 2015, pp. 3479--3487.
		
		\bibitem{gui2024survey}
		J.~Gui, T.~Chen, J.~Zhang, Q.~Cao, Z.~Sun, H.~Luo, D.~Tao, A survey on
		self-supervised learning: Algorithms, applications, and future trends, IEEE
		Transactions on Pattern Analysis and Machine Intelligence 46~(12) (2024)
		9052--9071.
		
		\bibitem{SimCLR}
		T.~Chen, S.~Kornblith, M.~Norouzi, G.~Hinton, A simple framework for
		contrastive learning of visual representations, in: Proceedings of the 37th
		International Conference on Machine Learning (ICML), 2020, pp. 1597--1607.
		
		\bibitem{chen2022data}
		N.~Chen, Z.~Xu, Z.~Liu, Y.~Chen, Y.~Miao, Q.~Li, Y.~Hou, L.~Wang, Data
		augmentation and intelligent recognition in pavement texture using a deep
		learning, IEEE Transactions on Intelligent Transportation Systems 23~(12)
		(2022) 25427--25436.
		
		\bibitem{May1976}
		R.~M. May, Simple mathematical models with very complicated dynamics, Nature
		261~(5560) (1976) 459--467.
		
		\bibitem{elkandoz2022image}
		M.~T. Elkandoz, W.~Alexan, Image encryption based on a combination of multiple
		chaotic maps, Multimedia Tools and Applications 81~(18) (2022) 25497--25518.
		
		\bibitem{Woo2023_ConvNeXtV2}
		S.~Woo, S.~Debnath, R.~Hu, X.~Chen, Z.~Liu, I.~S. Kweon, S.~Xie, Convnext v2:
		Co-designing and scaling convnets with masked autoencoders, in: Proceedings
		of the IEEE/CVF Conference on Computer Vision and Pattern Recognition (CVPR),
		2023, pp. 16133--16142.
		
		\bibitem{SENet}
		J.~Hu, L.~Shen, G.~Sun, Squeeze-and-excitation networks, in: Proceedings of the
		IEEE Conference on Computer Vision and Pattern Recognition (CVPR), 2018, pp.
		7132--7141.
		
		\bibitem{de2019classification}
		G.~V. de~Lima, P.~T. Saito, F.~M. Lopes, P.~H. Bugatti, Classification of
		texture based on bag-of-visual-words through complex networks, Expert Systems
		with Applications 133 (2019) 215--224.
		
		\bibitem{DeepTEN}
		H.~Zhang, J.~Xue, K.~Dana, Deep ten: Texture encoding network, in: Proceedings
		of the IEEE Conference on Computer Vision and Pattern Recognition (CVPR),
		2017, pp. 708--717.
		
		\bibitem{xu2021encoding}
		Y.~Xu, F.~Li, Z.~Chen, J.~Liang, Y.~Quan, Encoding spatial distribution of
		convolutional features for texture representation, Advances in Neural
		Information Processing Systems 34 (2021) 22732--22744.
		
		\bibitem{chen2021deep}
		Z.~Chen, F.~Li, Y.~Quan, Y.~Xu, H.~Ji, Deep texture recognition via exploiting
		cross-layer statistical self-similarity, in: Proceedings of the IEEE/CVF
		conference on Computer Vision and Pattern Recognition, 2021, pp. 5231--5240.
		
		\bibitem{tian2026hybrid}
		M.~Tian, L.~Tang, J.~Xu, Y.~Zhang, Y.~Yang, L.~Zeng, E.~Chen, Y.~Xie, Hybrid
		cnn-transformer framework with dynamic feature fusion for enhanced passport
		background texture classification: M. tian et al., The Visual Computer 42~(1)
		(2026) 4.
		
		\bibitem{he2020momentum}
		K.~He, H.~Fan, Y.~Wu, S.~Xie, R.~Girshick, Momentum contrast for unsupervised
		visual representation learning, in: Proceedings of the IEEE/CVF conference on
		computer vision and pattern recognition, 2020, pp. 9729--9738.
		
		\bibitem{he2022masked}
		K.~He, X.~Chen, S.~Xie, Y.~Li, P.~Doll{\'a}r, R.~Girshick, Masked autoencoders
		are scalable vision learners, in: Proceedings of the IEEE/CVF conference on
		computer vision and pattern recognition, 2022, pp. 16000--16009.
		
		\bibitem{liu2021self}
		X.~Liu, F.~Zhang, Z.~Hou, L.~Mian, Z.~Wang, J.~Zhang, J.~Tang, Self-supervised
		learning: Generative or contrastive, IEEE transactions on knowledge and data
		engineering 35~(1) (2021) 857--876.
		
		\bibitem{bafghi2025mixdiff}
		R.~A. Bafghi, N.~Harilal, C.~Monteleoni, M.~Raissi, Mix{D}iff: Mixing natural
		and synthetic images for robust self-supervised representations, in: 2025
		IEEE/CVF Winter Conference on Applications of Computer Vision (WACV), IEEE,
		2025, pp. 7500--7511.
		
		\bibitem{LU2025128393}
		J.~Lu, X.~Xia, X.~Zhang, R.~Zhao, Y.~Zhang,
		\href{https://www.sciencedirect.com/science/article/pii/S0957417425020123}{Multiple-image
			encryption algorithm based on a new 3d hyperchaotic map and whac-a-mole
			scrambling model}, Expert Systems with Applications 290 (2025) 128393.
		\newblock \href {https://doi.org/https://doi.org/10.1016/j.eswa.2025.128393}
		{\path{doi:https://doi.org/10.1016/j.eswa.2025.128393}}.
		\newline\urlprefix\url{https://www.sciencedirect.com/science/article/pii/S0957417425020123}
		
		\bibitem{jia2024generalized}
		B.~Jia, Z.~Guo, T.~Huang, F.~Guo, H.~Wu, A generalized lorenz system-based
		initialization method for deep neural networks, Applied Soft Computing 167
		(2024) 112316.
		
		\bibitem{sharan2014accuracy}
		L.~Sharan, R.~Rosenholtz, E.~H. Adelson, {Accuracy and speed of material
			categorization in real-world images}, Journal of Vision 14~(9) (2014) 12--12.
		\newblock \href {https://doi.org/10.1167/14.9.12} {\path{doi:10.1167/14.9.12}}.
		
		\bibitem{xu2009viewpoint}
		Y.~Xu, H.~Ji, C.~Ferm{\"u}ller, Viewpoint invariant texture description using
		fractal analysis, International Journal of Computer Vision 83~(1) (2009)
		85--100.
		\newblock \href {https://doi.org/10.1007/s11263-009-0220-6}
		{\path{doi:10.1007/s11263-009-0220-6}}.
		
		\bibitem{KTH}
		B.~Caputo, E.~Hayman, P.~Mallikarjuna, Class-specific material categorisation,
		in: Proceedings of the Tenth IEEE International Conference on Computer Vision
		(ICCV), Vol.~2, IEEE, 2005, pp. 1597--1604.
		
		\bibitem{CMB09}
		D.~Casanova, J.~J. de~Mesquita Sá~Junior, O.~M. Bruno, Plant leaf
		identification using gabor wavelets, International Journal of Imaging Systems
		and Technology 19~(3) (2009) 236--243.
		\newblock \href {https://doi.org/10.1002/ima.20201}
		{\path{doi:10.1002/ima.20201}}.
		
		\bibitem{DTD}
		M.~Cimpoi, S.~Maji, I.~Kokkinos, S.~Mohamed, A.~Vedaldi, Describing textures in
		the wild, in: Proceedings of the IEEE Conference on Computer Vision and
		Pattern Recognition (CVPR), 2014, pp. 3606--3613.
		
		\bibitem{xue2017differential}
		J.~Xue, H.~Zhang, K.~Dana, K.~Nishino, Differential angular imaging for
		material recognition, in: Proceedings of the IEEE Conference on Computer
		Vision and Pattern Recognition, 2017, pp. 764--773.
		
		\bibitem{mircea2016deep}
		C.~Mircea, M.~Subhransu, A.~Vedaldi, Deep filter banks for texture recognition,
		description, and segmentation, International Journal of Computer Vision
		(2016) 65--94.
		
		\bibitem{song2017locally}
		Y.~Song, F.~Zhang, Q.~Li, H.~Huang, L.~J. O'Donnell, W.~Cai,
		Locally-transferred fisher vectors for texture classification, in:
		Proceedings of the IEEE International Conference on Computer Vision, 2017,
		pp. 4912--4920.
		
		\bibitem{jbene2019fusion}
		M.~Jbene, A.~D. El~Maliani, M.~El~Hassouni, Fusion of convolutional neural
		network and statistical features for texture classification, in: 2019
		International Conference on Wireless Networks and Mobile Communications
		(WINCOM), IEEE, 2019, pp. 1--4.
		
		\bibitem{zhai2020deep}
		W.~Zhai, Y.~Cao, Z.-J. Zha, H.~Xie, F.~Wu, Deep structure-revealed network for
		texture recognition, in: Proceedings of the IEEE/CVF Conference on Computer
		Vision and Pattern Recognition, 2020, pp. 11010--11019.
		
		\bibitem{florindo2021visgraphnet}
		J.~B. Florindo, Y.-S. Lee, K.~Jun, G.~Jeon, M.~K. Albertini, Visgraphnet: A
		complex network interpretation of convolutional neural features, Information
		Sciences 543 (2021) 296--308.
		
		\bibitem{florindo2021using}
		J.~Florindo, K.~Metze, Using non-additive entropy to enhance convolutional
		neural features for texture recognition, Entropy 23~(10) (2021) 1259.
		
		\bibitem{mao2021deep}
		S.~Mao, D.~Rajan, L.~T. Chia, Deep residual pooling network for texture
		recognition, Pattern Recognition 112 (2021) 107817.
		
		\bibitem{yang2022dfaen}
		Z.~Yang, S.~Lai, X.~Hong, Y.~Shi, Y.~Cheng, C.~Qing, Dfaen: Double-order
		knowledge fusion and attentional encoding network for texture recognition,
		Expert Systems with Applications 209 (2022) 118223.
		
		\bibitem{scabini2023radam}
		L.~Scabini, K.~M. Zielinski, L.~C. Ribas, W.~N. Gon{\c{c}}alves, B.~De~Baets,
		O.~M. Bruno, Radam: Texture recognition through randomized aggregated
		encoding of deep activation maps, Pattern Recognition 143 (2023) 109802.
		
		\bibitem{mamidibathula2019texture}
		B.~Mamidibathula, S.~Amirneni, S.~S. Sistla, N.~Patnam, Texture classification
		using capsule networks, in: Iberian Conference on Pattern Recognition and
		Image Analysis, Springer, 2019, pp. 589--599.
		
		\bibitem{Lyra2024}
		L.~O. Lyra, A.~E. Fabris, J.~B. Florindo, A multilevel pooling scheme in
		convolutional neural networks for texture image recognition, Applied Soft
		Computing 152 (2024) 111282.
		\newblock \href {https://doi.org/10.1016/j.asoc.2023.111282}
		{\path{doi:10.1016/j.asoc.2023.111282}}.
		
		\bibitem{silva2021fractal}
		P.~M. Silva, J.~B. Florindo, Fractal measures of image local features: An
		application to texture recognition, Multimedia Tools and Applications 80~(9)
		(2021) 14213--14229.
		
		\bibitem{florindo2023boff}
		J.~B. Florindo, E.~E. Laureano, Boff: A bag of fuzzy deep features for texture
		recognition, Expert Systems with Applications 219 (2023) 119627.
		
	\end{thebibliography}
\end{document}